\documentclass{article}
\usepackage{colm2024_conference}
\pdfoutput=1
\usepackage{graphicx}
\usepackage{booktabs}
\usepackage{multirow}
\usepackage[table]{xcolor}
\usepackage[raster,skins]{tcolorbox}

\usepackage{amsmath,amsfonts,bm}

\def\eqref#1{equation~\ref{#1}}

\def\1{\bm{1}}

\DeclareMathAlphabet{\mathsfit}{\encodingdefault}{\sfdefault}{m}{sl}
\SetMathAlphabet{\mathsfit}{bold}{\encodingdefault}{\sfdefault}{bx}{n}

\title{FactorEngram: Factorized N-gram Memory with Basis-Level Gating for Language Models}
\author{%
{\small\bfseries Bowen Yang\textsuperscript{1,2,*}\quad
Jingbo Zhou\textsuperscript{1,$\dagger$}\quad
Qinghong Miao\textsuperscript{1}\quad
Hua Wu\textsuperscript{1,$\dagger$}}\\[0.9em]
{\small\textsuperscript{1}Large Model Frontier Research Department, Baidu Inc., China}\\
{\small\textsuperscript{2}Nanyang Technological University}\\[0.6em]
{\small\nolinkurl{{yangbowen06,zhoujingbo,miaoqinghong,wu_hua}@baidu.com}}%
}

\makeatletter
\long\def\@abstract{
Lookup-based memory has been a promising way to scale the parameters of large
language models (LLMs). It retrieves learned representations of local token
patterns, such as $n$-grams, instead of reconstructing them through
successive layers of computation. However, existing designs such as Engram
treat each retrieved embedding as a \emph{monolithic} unit. Each embedding
is stored in its own hashed slot and modulated by a single scalar gate. As
a result, polysemous patterns cannot selectively read out the components of
their memory that are relevant to the context. Moreover, parameters are
shared only through hash collisions, which are largely unrelated to
semantics. We propose \textbf{FactorEngram}, a factorized $n$-gram memory
with basis-level contextual gating. FactorEngram retrieves
sparsity-regularized coefficients over a dictionary of basis vectors shared
across patterns, so related patterns can reuse common components. The same
dictionary is also used for gating. The backbone hidden state is scored
against each basis vector to gate the corresponding coefficient before
reconstruction, which lets the context modulate each memory component
individually. FactorEngram also covers both individual tokens and
multi-token $n$-grams, and we systematically study where the memory branch
should be inserted. On 340M- and 1B-parameter Transformer backbones,
FactorEngram improves language modeling and downstream task performance. Ablation studies confirm the contribution
of each component and identify insertion before the attention sublayer in
the middle layers as an effective configuration. 

}
\makeatother

\begin{document}
\maketitle
\renewcommand{\thefootnote}{\fnsymbol{footnote}}
\footnotetext[1]{This work was done when the first author was an intern in Baidu Inc., under the supervision of Jingbo Zhou.}
\footnotetext[2]{Jingbo Zhou and Hua Wu are corresponding authors.}
\renewcommand{\thefootnote}{\arabic{footnote}}


\section{Introduction}
\label{sec:introduction}

Lookup-based memory has emerged as a promising direction for scaling the
parameters of large language models (LLMs). This direction is motivated by the observation that many
recurring expressions, such as ``the Eiffel Tower'' and ``Mount Everest,''
are associated with relatively static lexical and factual knowledge. Yet
standard Transformers ~\citep{vaswaniAttentionAllYou2017} must reconstruct the representations of such
expressions through successive layers of computation. Lookup-based memory
addresses this inefficiency by augmenting the backbone with an auxiliary
memory branch that retrieves learned representations of local token
patterns, providing direct access to pattern-specific information instead
of relying entirely on the backbone to reconstruct
it~\citep{chengConditionalMemoryScalable2026}.

A typical lookup-based memory module maps local patterns, such as
$n$-grams, to table addresses via direct indexing or hashing, and retrieves
the corresponding learnable embeddings. These embeddings are then
aggregated and incorporated into the backbone computation, optionally after
being modulated by the current hidden state. Recent architectures,
including Gemma~3n's Per-Layer Embeddings~\citep{google2025gemma3nple},
Engram~\citep{chengConditionalMemoryScalable2026},
STEM~\citep{sadhukhanSTEMScalingTransformers2026}, and
LongCat-Flash-Lite~\citep{liuScalingEmbeddingsOutperforms2026}, explore
different instantiations of such modules. Among them, Engram offers a
representative implementation of conditional $n$-gram memory and has been
adopted in DeepSeek-V4.1-Flash~\citep{xu2026deepseekv4.1}, demonstrating its
applicability at large scale.

Despite this progress, we observe that Engram treats each retrieved
$n$-gram embedding as a \emph{monolithic} unit: each embedding occupies an
independent hashed slot and is injected into the backbone through a single
scalar gate. This monolithic design gives rise to two limitations.
First, \textbf{polysemous patterns cannot be selectively read out according
to context.} The same pattern may call for different semantic components of
its memory in different contexts; for example, ``the bank'' may refer to a
financial institution or to the land alongside a river. A scalar gate can
only amplify or suppress the embedding as a whole, and therefore cannot
retain the context-relevant components while suppressing the irrelevant
ones. Second, \textbf{parameter sharing is unrelated to semantics.} Because
the space of $n$-gram combinations is prohibitively large, the memory
relies on hashing to map $n$-grams to table entries. Consequently,
parameters are shared only among $n$-grams that collide under the hash
functions, which are typically semantically unrelated, whereas semantically
similar $n$-grams have no mechanism to share any part of their
representations. Both limitations stem from a common cause: the smallest
unit of memory is an $n$-gram embedding vector. This calls for a joint design of memory representation and contextual modulation, in which shared semantic components can be reused across patterns and individually adapted to each context.

\begin{figure*}[t]
\vspace{-10pt}
\centering
\includegraphics[width=0.82\textwidth]{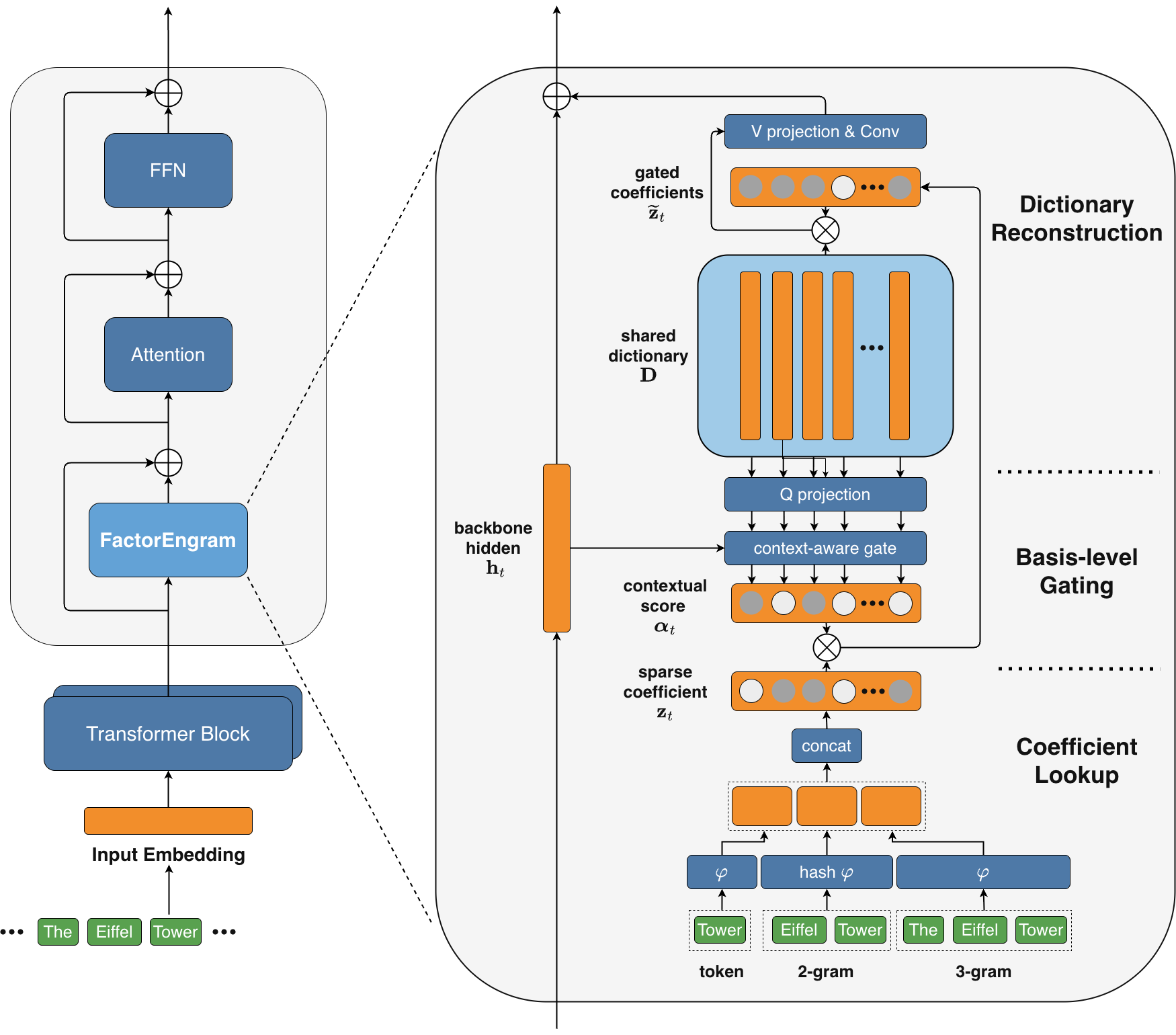}
\caption{Overview of FactorEngram. Left: memory modules are residually inserted
before the attention module at selected Transformer layers. Right: n-gram lookups retrieve coefficients that are concatenated
and modulated by context-dependent basis-level gates. The shared
dictionary participates in both gating and memory reconstruction.
The reconstructed memory is projected, refined by a short causal
convolution, and added to the backbone hidden state.}
\label{fig:FactorEngram-overview}
\vspace{-10pt}
\end{figure*}

To this end, we propose \textbf{FactorEngram}, a factorized $n$-gram memory
architecture with basis-level contextual gating as illustrated in Figure~\ref{fig:FactorEngram-overview}. Rather than retrieving
complete memory embeddings, FactorEngram retrieves learnable coefficients
over a dictionary shared across all local patterns within each memory
module. Specifically, multiple lookup heads retrieve the coefficients of
each $n$-gram from hashed tables and concatenate them, so that each
coefficient corresponds to one dictionary basis vector. The coefficient
tables store pattern-specific information, whereas the dictionary provides
a set of shared basis vectors whose linear combination reconstructs the
memory content. Different patterns therefore reuse the same basis vectors
with different coefficients, allowing related patterns to share parameters
through common basis vectors rather than through hash collisions. Drawing on sparse coding~\citep{olshausenEmergenceSimplecellReceptive1996} and its
applications to language model
representations~\citep{brickenMonosemanticityDecomposingLanguage2023,templeton2024scaling},
we further impose an $\ell_1$ penalty on the retrieved coefficients,
encouraging each pattern to rely on a small subset of basis vectors. 

Crucially, FactorEngram uses the same dictionary for both contextual gating
and memory reconstruction. The current backbone hidden state is projected
into a query, which is scored against each basis vector to produce a gate
for the corresponding coefficient. These gates scale the retrieved
coefficients before reconstruction, allowing the context to determine the
contribution of each memory component individually. The vectors used to
assess contextual relevance are thus exactly those used to reconstruct the
output. The reconstructed memory is then projected to the backbone width,
refined by a short causal convolution, and added to the residual stream,
leaving the backbone's attention and feed-forward modules unchanged. The
memory parameters and the backbone are trained jointly with the
language-modeling objective and the coefficient sparsity penalty.

Beyond factorization and gating, FactorEngram further refines two design choices
in existing memory modules. First, existing methods cover only a subset of
local patterns: STEM retrieves embeddings only for individual tokens,
whereas Engram retrieves only 2-grams and 3-grams. FactorEngram covers both
individual tokens and multi-token $n$-grams. Second, existing methods
insert the memory branch at a fixed set of positions. We systematically
study where the memory branch should be inserted, both across layers and
relative to the attention and feed-forward sublayers.

Experiments with Transformer backbones of 340M and 1B parameters show that
FactorEngram improves language modeling, downstream task performance, and
long-context retrieval. Ablation studies quantify the contribution of each
component and the effect of sparsity regularization, and placement
experiments identify insertion before the attention sublayer in the middle
layers as an effective configuration.

Our contributions are summarized as follows:
\begin{itemize}
    \item We introduce FactorEngram, a factorized $n$-gram memory
    architecture that represents local token patterns using
    sparsity-regularized coefficients over a shared dictionary, enabling
    related patterns to share components rather than relying on hash
    collisions.
    \item We design basis-level contextual gating, which reuses the
    reconstruction dictionary to assess contextual relevance and to
    modulate each memory component independently before reconstruction.
    \item We evaluate FactorEngram at two backbone scales, demonstrating
    gains in language modeling, downstream accuracy, and long-context
    retrieval, and investigate its architectural components, sparsity
    regularization, pattern coverage, and memory placement through
    controlled studies.
\end{itemize}


\section{Preliminaries}
\label{sec:preliminaries}

In this section, we define learnable lookup-based memory for LLMs as a module that contains memory table whose entries are learnable and addressed by local token patterns and incorporates retrieved memory contents into backbone computation. We formulate its general architecture by describing it with five operations:
discrete addressing, table lookup and branch aggregation,
contextual modulation, memory output mapping, and backbone
integration.

\textbf{Discrete addressing.}
Given a token sequence $X=(x_1,\ldots,x_T)$ from a vocabulary
$\mathcal{V}$, the memory module is inserted at each position $t$ in specific insertion layers $\ell\in\mathcal{I}$.
Each retrieval branch $b\in\mathcal{B}_{\ell}$ is associated with
a suffix length $n_b$ and an address space containing
$M_b^{(\ell)}$ entries.
Its input is the suffix
$g_{t,n_b}=(x_{t-n_b+1},\ldots,x_t)\in\mathcal{V}^{n_b}$.
The addressing function is defined as
\begin{equation}
\phi_b^{(\ell)}:
\mathcal{V}^{n_b}
\rightarrow
\{0,\ldots,M_b^{(\ell)}-1\},
\qquad
i_{t,b}^{(\ell)}
=
\phi_b^{(\ell)}(g_{t,n_b})
\label{eq:memory-addressing}
\end{equation}
The case $n_b=1$ corresponds to unigrams and
$n_b>1$ corresponds to longer N-gram patterns.
Multiple branches may use the same suffix length, allowing
different hash heads.
Importantly, the table address depends on the local discrete input
sequence rather than backbone hidden states.

\textbf{Table lookup and branch aggregation.}
Each branch retrieves an entry from a learnable table
$\mathbf{E}_b^{(\ell)}
\in\mathbb{R}^{M_b^{(\ell)}\times c_b^{(\ell)}}$,
where $c_b^{(\ell)}$ is the dimension of each table entry.
The retrieved entries from all branches are then aggregated:
\begin{equation}
\mathbf{u}_{t,b}^{(\ell)}
=
\mathbf{E}_b^{(\ell)}[i_{t,b}^{(\ell)}],
\qquad
\mathbf{z}_t^{(\ell)}
=
\mathcal{A}_{\ell}
\left(
\{\mathbf{u}_{t,b}^{(\ell)}\}_{b\in\mathcal{B}_{\ell}}
\right)
\label{eq:memory-lookup}
\end{equation}
The aggregation function $\mathcal{A}_{\ell}$ may be concatenation,
summation, or other combinations. Tables may be
layer-specific or shared across layers.

\textbf{Contextual modulation.}
Let $\mathbf{h}_t^{(\ell)}\in\mathbb{R}^{d}$ denote the backbone
hidden state at layer $\ell$ and position $t$, which compresses information from history context.
Contextual modulation adjusts the retrieved representation using this state:
\begin{equation}
\widetilde{\mathbf{z}}_t^{(\ell)}
=
\mathcal{C}_{\ell}
\left(
\mathbf{h}_t^{(\ell)},\mathbf{z}_t^{(\ell)};
\Theta_{\ell}
\right)
\label{eq:memory-modulation}
\end{equation}
Here, $\Theta_{\ell}$ denotes memory-module parameters, which may
be shared across operations. In architectures without contextual
modulation design, $\mathcal{C}_{\ell}$ performs as identity.

\textbf{Memory output mapping.}
An output mapping converts the modulated representation into
a memory contribution for backbone integration:
\begin{equation}
\mathbf{m}_t^{(\ell)}
=
\mathcal{R}_{\ell}
\left(
\widetilde{\mathbf{z}}_{\le t}^{(\ell)};
\Theta_{\ell}
\right)
\label{eq:memory-output}
\end{equation}
The notation $\widetilde{\mathbf{z}}_{\le t}^{(\ell)}$
denotes the representations up to position $t$, allowing causal
operations such as a short convolution.

\textbf{Backbone integration.}
An integration function specifies how the memory contribution
enters backbone computation:
\begin{equation}
\widehat{\mathbf{h}}_t^{(\ell)}
=
\mathcal{F}_{\ell}
\left(
\mathbf{h}_t^{(\ell)},\mathbf{m}_t^{(\ell)}
\right)
\label{eq:memory-integration}
\end{equation}
For residual integration,
$\widehat{\mathbf{h}}_t^{(\ell)}
=\mathbf{h}_t^{(\ell)}+\mathbf{m}_t^{(\ell)}$.
The fused state then enters the subsequent backbone computation.
Memory may insert in a feed-forward sublayer or augment
the input embeddings; the latter is treated as an integration
stage before the first block. The next section instantiates
these operations for our framework FactorEngram.

\section{Method}
\label{sec:method}

Figure~\ref{fig:FactorEngram-overview} illustrates FactorEngram architecture. It instantiates the lookup-based memory in
Section~\ref{sec:preliminaries} by retrieving sparsity-regularized
coefficients over a dictionary shared across local patterns and applying
basis-level contextual gating (Figure~\ref{fig:FactorEngram-overview}).
The gated coefficients reconstruct a memory vector, which is refined by a
short causal convolution and added to the backbone. Each insertion layer
has separate memory parameters; we omit layer indices below unless needed.

\subsection{Memory Retrieval: Sparse Dictionary Coefficients}
\label{sec:factorized-memory}

\textbf{Coefficient representation.}
FactorEngram stores pattern as linear coefficients over a shared
dictionary $\mathbf{D}\in\mathbb{R}^{s\times d_m}$, instead of directly assigning an independent dense embedding to every local pattern. $s$ is the total
coefficient width and hence the number of dictionary basis vectors,
and $d_m$ is the dimension of the memory representation space.
We refer to the dictionary row vectors as basis vectors,
without requiring linear independence. $\mathbf{d}_j^{\top}$ is the $j$-th row of the dictionary that defines a basis vector in the
$d_m$-dimensional memory space, and the $j$-th retrieved coefficient
is its corresponding weight. The dictionary is shared across different patterns
within a layer, while coefficients are stored in pattern-addressed
tables. Both the coefficients
and dictionary are learnable. The factorization permits the representation of different patterns to share the same basis
vector parameters with different coefficient assignments. An $L_1$ penalty encourages the sparsity of coefficients, as defined in Section~\ref{sec:FactorEngram-training}.

\textbf{Addressing and aggregation.}
We retrieve local patterns at unigram, bigram, and trigram
granularities using multiple lookup heads per suffix length.
Each head corresponds to a branch $b\in\mathcal{B}_{\ell}$ in
Section~\ref{sec:preliminaries}, with suffix length
$n_b\in\{1,2,3\}$. Each branch has its own deterministic mapping
function $\phi_b$ and learnable coefficient table
$\mathbf{E}_b\in\mathbb{R}^{M_b\times s_b}$, where $s_b$ is
its entry width. We define $\phi_b$ as direct indexing when $n_b=1$ and as hash function when $n_b\geq 2$. Given the suffix $g_{t,n_b}$, we retrieve
and concatenate the branch coefficients:
\begin{equation}
\mathbf{z}_{t,b}
=\mathbf{E}_b[\phi_b(g_{t,n_b})]\in\mathbb{R}^{s_b}
\qquad
\mathbf{z}_t
=\operatorname{Concat}_{b\in\mathcal{B}_{\ell}}
\bigl(\mathbf{z}_{t,b}\bigr)\in\mathbb{R}^{s}
\label{eq:FactorEngram-retrieval}
\end{equation}
where $s=\sum_{b\in\mathcal{B}_{\ell}}s_b$.
Concatenation follows a fixed branch order, implements
$\mathcal{A}_{\ell}$, and aligns the retrieved coordinates
with the dictionary rows. The resulting $\mathbf{z}_t$ is
passed to contextual modulation before dictionary reconstruction.

\subsection{Contextual Modulation: Basis-Level Gating}
\label{sec:basis-gating}

The same local pattern can call for different memory contents
in different contexts. Therefore, FactorEngram modulates each
dictionary coefficient separately before combining the basis
vectors. We project the current backbone state $\mathbf{h}_t$ into memory space
using a query projection $\mathbf{Q}\in\mathbb{R}^{d\times d_m}$, normalize the
projected query using RMSNorm~\citep{zhangRootMeanSquare2019}, and compute its dot-product scores against dictionary rows:
\begin{equation}
\mathbf{q}_t
=\operatorname{RMSNorm}(\mathbf{Q}^{\top}\mathbf{h}_t),
\qquad
\boldsymbol{\alpha}_t
=\sigma\!\left(\frac{\mathbf{D}\mathbf{q}_t}{\sqrt{d}}\right)
\in(0,1)^s
\label{eq:FactorEngram-basis-gate}
\end{equation}
Here, $d$ is the backbone hidden width and $\sigma$ is sigmoid. Each weight scales the corresponding
retrieved coefficient:
\begin{equation}
\widetilde{\mathbf{z}}_t
=\mathbf{z}_t\odot\boldsymbol{\alpha}_t
\label{eq:FactorEngram-modulation}
\end{equation}
Thus,
$\boldsymbol{\alpha}_t$ controls the memory representation according to context. This basis-level gating instantiates the contextual
modulation function $\mathcal{C}_{\ell}$.

\subsection{Memory Output and Integration: Dictionary Reconstruction}
\label{sec:memory-integration}

\textbf{Memory output mapping.}
The modulated coefficients correspond to a coordinate in the linear space of shared
dictionary. We reconstruct the memory vector through linear combination and project
it to the backbone width using $\mathbf{V}\in\mathbb{R}^{d\times d_m}$:
\begin{equation}
\mathbf{e}_t
=\mathbf{D}^{\top}\widetilde{\mathbf{z}}_t
=\sum_{j=1}^{s}\widetilde z_{t,j}\mathbf{d}_j,
\qquad
\mathbf{v}_t=\mathbf{V}\mathbf{e}_t
\label{eq:FactorEngram-reconstruction}
\end{equation}
Following Engram~\citep{chengConditionalMemoryScalable2026}, we apply a short depthwise causal convolution to combine
memory outputs from neighboring positions, with a residual
connection preserving the current-position output:
\begin{equation}
\mathbf{m}_t
=\mathbf{v}_t
+\operatorname{SiLU}\!\left(
\operatorname{Conv1D}\!\left(
\operatorname{RMSNorm}(\mathbf{v}_{\le t})
\right)_t\right)
\label{eq:FactorEngram-convolution}
\end{equation}
Dictionary reconstruction, value projection, and convolution
together implement $\mathcal{R}_{\ell}$, mapping
$\widetilde{\mathbf{z}}_{\le t}$ to $\mathbf{m}_t$ without using
future positions.

\textbf{Residual integration.}
FactorEngram uses the memory module as an auxiliary branch while preserving
the backbone computation. It adds the memory output to the
current backbone state:
\begin{equation}
\widehat{\mathbf{h}}_t
=\mathcal{F}_{\ell}(\mathbf{h}_t,\mathbf{m}_t)
=\mathbf{h}_t+\mathbf{m}_t
\label{eq:FactorEngram-residual}
\end{equation}
In the default configuration, this update precedes attention module at
the selected insertion layers. The fused states continue through
the backbone attention and feed-forward modules, which remain
intact. We evaluate insertion depth and alternative integration
locations in the experiments.

\subsection{Training Objective: Sparsity Regularization}
\label{sec:FactorEngram-training}

\textbf{Coefficient sparsity.}
Following the principles of sparse coding and its applications to language
model representations~\citep{olshausenEmergenceSimplecellReceptive1996,
brickenMonosemanticityDecomposingLanguage2023,
templeton2024scaling},
we regularize the retrieved representations by encouraging
each local pattern to rely on a small subset of dictionary
components. We apply an $L_1$ penalty to the concatenated coefficients
$\mathbf{z}_t$ before contextual modulation. At each position,
the penalty is averaged over the set of memory insertion layers $\mathcal{I}$:
\begin{equation}
\mathcal{L}_{\mathrm{sparsity}}
=\frac{1}{T}
\sum_{t=1}^T
\frac{1}{|\mathcal{I}|}
\sum_{\ell\in\mathcal{I}}
\left\|\mathbf{z}_t^{(\ell)}\right\|_1
\label{eq:FactorEngram-sparsity}
\end{equation}

\textbf{Joint optimization.}
We train the memory module and backbone jointly with the
next-token prediction objective:
\begin{equation}
\mathcal{L}_{\mathrm{pretrain}}
=\mathcal{L}_{\mathrm{NLL}}
+\lambda\mathcal{L}_{\mathrm{sparsity}},
\label{eq:FactorEngram-objective}
\end{equation}
where $\lambda$ controls the sparsity regularization strength.
The coefficient tables, dictionary, projections, and convolution
are learned together with the backbone, without predefined
semantic labels for the basis vectors.

\section{Experiments}
\label{sec:experiments}

We evaluate FactorEngram through comparisons at two backbone scales,
component ablations, and memory-placement studies. The experiments
measure language modeling, downstream accuracy, and long-context
retrieval, and examine how these outcomes depend on basis-level
gating, unigram retrieval, sparsity regularization, and insertion
configuration.

\subsection{Experimental Setup}
\label{sec:experimental-setup}

\textbf{Models and training.}
We jointly train FactorEngram with 340M-parameter and 1B-parameter Transformer~\citep{vaswaniAttentionAllYou2017} backbones from scratch using 30B and
120B tokens from FineWeb-Edu~\citep{HuggingFaceFWFinewebeduDatasets}, respectively, with a maximum context length of 8192 (8K). The token vocabulary size is 32K.
The 340M backbone configuration uses 1B lookup-table parameters,
with both bigram and trigram table capacities of 250K entries. The 1B backbone configuration uses 2B lookup-table parameters, with
the bigram and trigram table capacities of 250K and 750K entries,
respectively. Unless otherwise specified, the memory modules are inserted before attention at layers 10 and 12, with sparsity strength 
$\lambda=10^{-3}$ defined in Eq.~\ref{eq:FactorEngram-objective}. Configuration details are listed in Appendix~\ref{app:model-configurations}.

\textbf{Evaluation.}
We compare FactorEngram with pure Transformer backbones at both scales
and with our reproduction of Engram~\citep{chengConditionalMemoryScalable2026} at both scales. The former
comparison assesses the contribution of sparsely-activated lookup memory; the latter assesses the contribution of factorization for memory representation and modulation.

We evaluate models on language modeling, downstream accuracy, and long-context retrieval. Language modeling is evaluated by perplexity on WikiText~\citep{merityPointerSentinelMixture2016} and
LAMBADA~\citep{papernoLAMBADADatasetWord2016} (OpenAI variant). Downstream accuracy (\%) are assessed on PIQA~\citep{biskPIQAReasoningPhysical2020}, HellaSwag~\citep{zellersHellaSwagCanMachine2019},
WinoGrande~\citep{sakaguchiWinoGrandeAdversarialWinograd2021}, ARC-Easy and ARC-Challenge~\citep{clarkThinkYouHave2018}, Social IQA~\citep{sapSocialIQaCommonsense2019}, and BoolQ~\citep{clarkBoolQExploringSurprising2019}.
Long-context retrieval is evaluated by accuracy (\%) in three 8K-context-length
Needle-in-a-Haystack~\citep{gkamradtGkamradtLLMTest_NeedleInAHaystack2026} variants of increasing difficulty, denoted as
NIAH-1, NIAH-2, and NIAH-3.

\subsection{Main Results}
\label{sec:main-results}

Table~\ref{tab:main-results} with detailed version in Appendix~\ref{app:main-result-detail} compares FactorEngram with the
baselines at two backbone scales. FactorEngram improves overall evaluation performance at both
scales, with particularly strong gains in long-context retrieval.
At 340M, FactorEngram reduces WikiText perplexity from 23.19
to 21.29 and LAMBADA perplexity from 24.90 to 21.69,
while improving average downstream accuracy by
1.76 percentage points over the Transformer.
It also outperforms Engram on all evaluation metrics,
with the largest accuracy gains on the harder retrieval
tasks: 41.5 and 43.3 percentage points on NIAH-2 and NIAH-3,
respectively.
Extending FactorEngram to the 1B backbone preserves these improvements on most metrics, with WikiText perplexity comparable to the Transformer and slightly lower downstream accuracy than Engram.
The retrieval gains remain substantial at this scale,
ranging from 9.3 to 27.6 percentage points over Engram.

\begin{table*}[t]
\centering
\caption{Main results with an 8K context length. The 340M and 1B
backbones are trained on 30B and 120B tokens, respectively. Accuracy
metrics are reported as percentages. Bold indicates the best value
within each model scale.}
\label{tab:main-results}
\resizebox{0.9\textwidth}{!}{%
\begin{tabular}{l|cc|c|ccc}
\toprule
\multirow{2}{*}{Model}
& \multicolumn{2}{c|}{Language Modeling PPL}
& \multicolumn{1}{c|}{Downstream}
& \multicolumn{3}{c}{Long-context Retrieval} \\
\cmidrule(lr){2-3}
\cmidrule(lr){4-4}
\cmidrule(lr){5-7}
& WikiText $\downarrow$
& LAMBADA $\downarrow$
& Avg. Acc. $\uparrow$
& NIAH-1 $\uparrow$
& NIAH-2 $\uparrow$
& NIAH-3 $\uparrow$ \\
\midrule
\rowcolor{gray!15}
\multicolumn{7}{c}{
\rule[-0.7ex]{0pt}{3.0ex}\textit{340M backbone}
} \\
Transformer & 23.19 & 24.90 & 48.10 & 44.2 & 47.7 & 22.1 \\
Engram & 22.29 & 23.94 & 49.06 & 69.8 & 37.6 & 15.0 \\
FactorEngram & \textbf{21.29} & \textbf{21.69}
& \textbf{49.82} & \textbf{70.6} & \textbf{79.1} & \textbf{58.3} \\
\midrule
\rowcolor{gray!15}
\multicolumn{7}{c}{
\rule[-0.7ex]{0pt}{3.0ex}\textit{1B backbone}
} \\
Transformer & \textbf{15.57} & 11.73 & 55.61
& 47.5 & 72.8 & 25.7 \\
Engram & 15.60 & 11.83 & \textbf{55.81}
& 45.8 & 75.2 & 16.6 \\
FactorEngram & \textbf{15.57} & \textbf{11.39}
& 55.80 & \textbf{72.0} & \textbf{84.5} & \textbf{44.2} \\
\bottomrule
\end{tabular}%
}
\end{table*}

\subsection{Ablation Studies}
\label{sec:ablation-studies}

We conduct ablations with the 340M backbone to examine
the contributions of factorized memory, basis-level gating,
unigram retrieval, and sparsity regularization.
To assess basis-level gating, we replace it with a scalar gating while retaining
the coefficient-dictionary representation. This scalar-gating variant first goes through dictionary reconstruction with $\mathbf{e}_t=\mathbf{D}^{\top}\mathbf{z}_t$ modified from Eq.~\ref{eq:FactorEngram-reconstruction}, and then applies gating by scaling $\mathbf{e}$ with the scalar dot-product score between $\mathbf{q}_t$ and $\mathbf{e}_t$:

\begin{equation}
\alpha_t^{\mathrm{SCALAR}}
=\sigma\!\left(
\frac{\mathbf{e}_t^\top\mathbf{q}_t}{\sqrt{d}}
\right)
\in(0,1),
\qquad
\widetilde{\mathbf{e}}_t
= \alpha_t^{\mathrm{SCALAR}}\mathbf{e}_t
\label{eq:exp-ablation-scalar-gate}
\end{equation}

This implements a scalar gating module same as the Engram gating module while preserving our factorization representation before gating.
We further assess the factorization design as a whole by jointly
removing the coefficient-dictionary representation and
basis-level gating. Our reproductions of
Engram and Engram with unigram retrieval serve as such factorization
ablations for FactorEngram. In addition to scalar gating in Eq.~\ref{eq:exp-ablation-scalar-gate}, its lookup table is modified to directly store embedding $\mathbf{e}_t$ in each row rather than represented by $\mathbf{z}_t$ and $\mathbf{D}$.
We also remove unigram retrieval in FactorEngram
while retaining the factorization design.
Table~\ref{tab:component-ablation} reports these module ablation results.
For sparsity, we vary sparsity penalty strength $\lambda$ in Eq.~\ref{eq:FactorEngram-objective} to assess
the effect of sparsity regularization on coefficients.

\textbf{Factorized memory.}
The complete factorization design improves overall performance
over dense lookup memory of Engram in both settings with and without unigrams.
Compared with Engram (+ unigram), FactorEngram
reduces WikiText perplexity from 22.32 to 21.29 and LAMBADA
perplexity from 24.42 to 21.69, while increasing average
accuracy by 0.92 percentage points.
The largest gains occur on harder NIAH-2 and NIAH-3, improving
by 36.9 and 29.1 percentage points, respectively.
These results support the joint coefficient representation
and basis-level gating design.

\textbf{Basis-level gating.}
Basis-level gating is important for realizing full benefits
of the factorized representation.
Replacing it with scalar gating degrades performance
across all evaluation metrics, with
average downstream accuracy dropping from 49.82 to 48.15 and the hardest  NIAH-3
from 58.3 to 9.4.
These results support modulating individual dictionary
coefficients rather than uniformly scaling the reconstructed
memory vector.

\textbf{Unigram retrieval.}
Unigram retrieval complements multi-token memory in FactorEngram.
Removing it degrades all evaluation metrics, with
particularly large drops on NIAH-2 and NIAH-3.
Adding unigram retrieval to Engram also improves all three
NIAH scores, although its perplexity and average accuracy
slightly worsen.
Unigram retrieval improves long-context retrieval in both
architectures, while also improving language modeling
and downstream accuracy in FactorEngram.

\textbf{Coefficient sparsity.}
We vary $\lambda\in\{0,10^{-4},10^{-3},10^{-2}\}$ to assess
the effect of coefficient sparsity regularization
(Table~\ref{tab:sparsity-results}).
Overall performance generally improves as $\lambda$
increases from 0 to $10^{-3}$, but declines at $10^{-2}$.
Although FactorEngram already outperforms the Transformer
across all evaluation metrics without sparsity regularization,
$\lambda=10^{-3}$ performs best on all evaluation metrics
except WikiText perplexity, where $\lambda=10^{-4}$
achieves the lowest value.
Increasing $\lambda$ to $10^{-2}$ makes all evaluation
metrics worse than those of the non-regularized model,
supporting the choice of a moderate regularization strength.

\begin{table*}[t]
\centering
\caption{Architectural component ablations with the 340M backbone.
All models use an 8K context length. Accuracy metrics are reported
as percentages.
Bold indicates the best value in each column.}
\label{tab:component-ablation}
\resizebox{0.9\textwidth}{!}{%
\begin{tabular}{l|cc|c|ccc}
\toprule
\multirow{2}{*}{Configuration}
& \multicolumn{2}{c|}{Language Modeling PPL}
& \multicolumn{1}{c|}{Downstream}
& \multicolumn{3}{c}{Long-context Retrieval} \\
\cmidrule(lr){2-3}
\cmidrule(lr){4-4}
\cmidrule(lr){5-7}
& WikiText $\downarrow$
& LAMBADA $\downarrow$
& Avg. Acc. $\uparrow$
& NIAH-1 $\uparrow$
& NIAH-2 $\uparrow$
& NIAH-3 $\uparrow$ \\
\midrule
Transformer & 23.19 & 24.90 & 48.10 & 44.2 & 47.7 & 22.1 \\
Engram & 22.29 & 23.94 & 49.06 & 69.8 & 37.6 & 15.0 \\
Engram (+ unigram) & 22.32 & 24.42 & 48.94
& \textbf{74.0} & 42.2 & 29.2 \\
FactorEngram (scalar gate) & 22.85 & 25.66 & 48.15
& 67.8 & 45.2 & 9.4 \\
FactorEngram (- unigram) & 22.74 & 23.76 & 49.20
& 60.2 & 44.2 & 23.0 \\
\rowcolor{gray!15}
FactorEngram & \textbf{21.29} & \textbf{21.69}
& \textbf{49.82} & 70.6 & \textbf{79.1} & \textbf{58.3} \\
\bottomrule
\end{tabular}%
}
\end{table*}

\begin{table*}[t]
\centering
\caption{Effect of the sparsity coefficient $\lambda$ with the 340M
backbone and an 8K context length. Accuracy metrics are reported as
percentages. Bold indicates the best value in each column.}
\label{tab:sparsity-results}
\resizebox{0.9\textwidth}{!}{%
\begin{tabular}{l|cc|c|ccc}
\toprule
\multirow{2}{*}{Configuration}
& \multicolumn{2}{c|}{Language Modeling PPL}
& \multicolumn{1}{c|}{Downstream}
& \multicolumn{3}{c}{Long-context Retrieval} \\
\cmidrule(lr){2-3}
\cmidrule(lr){4-4}
\cmidrule(lr){5-7}
& WikiText $\downarrow$
& LAMBADA $\downarrow$
& Avg. Acc. $\uparrow$
& NIAH-1 $\uparrow$
& NIAH-2 $\uparrow$
& NIAH-3 $\uparrow$ \\
\midrule
Transformer & 23.19 & 24.90 & 48.10 & 44.2 & 47.7 & 22.1 \\
$\lambda=0$ & 21.47 & 22.58 & 49.49 & 58.8 & 70.6 & 31.8 \\
$\lambda=10^{-4}$ & \textbf{21.18} & 22.43 & 49.43
& 67.4 & 74.0 & 43.2 \\
\rowcolor{gray!15}
$\lambda=10^{-3}$ & 21.29 & \textbf{21.69}
& \textbf{49.82} & \textbf{70.6} & \textbf{79.1} & \textbf{58.3} \\
$\lambda=10^{-2}$ & 23.89 & 25.79 & 47.77 & 50.0 & 50.0 & 17.8 \\
\bottomrule
\end{tabular}%
}
\end{table*}

\subsection{Memory Placement}
\label{sec:memory-placement}

We examine memory placement from coarse to fine using
the 24-layer, 340M backbone.
We first sweep insertion depths across the backbone,
then refine placement within the selected layers by
comparing different insertion points.

\subsubsection{Insertion Depth}
\label{sec:insertion-layers}

\begin{figure*}[t]
\vspace{-10pt}
\centering
\includegraphics[width=0.9\textwidth]{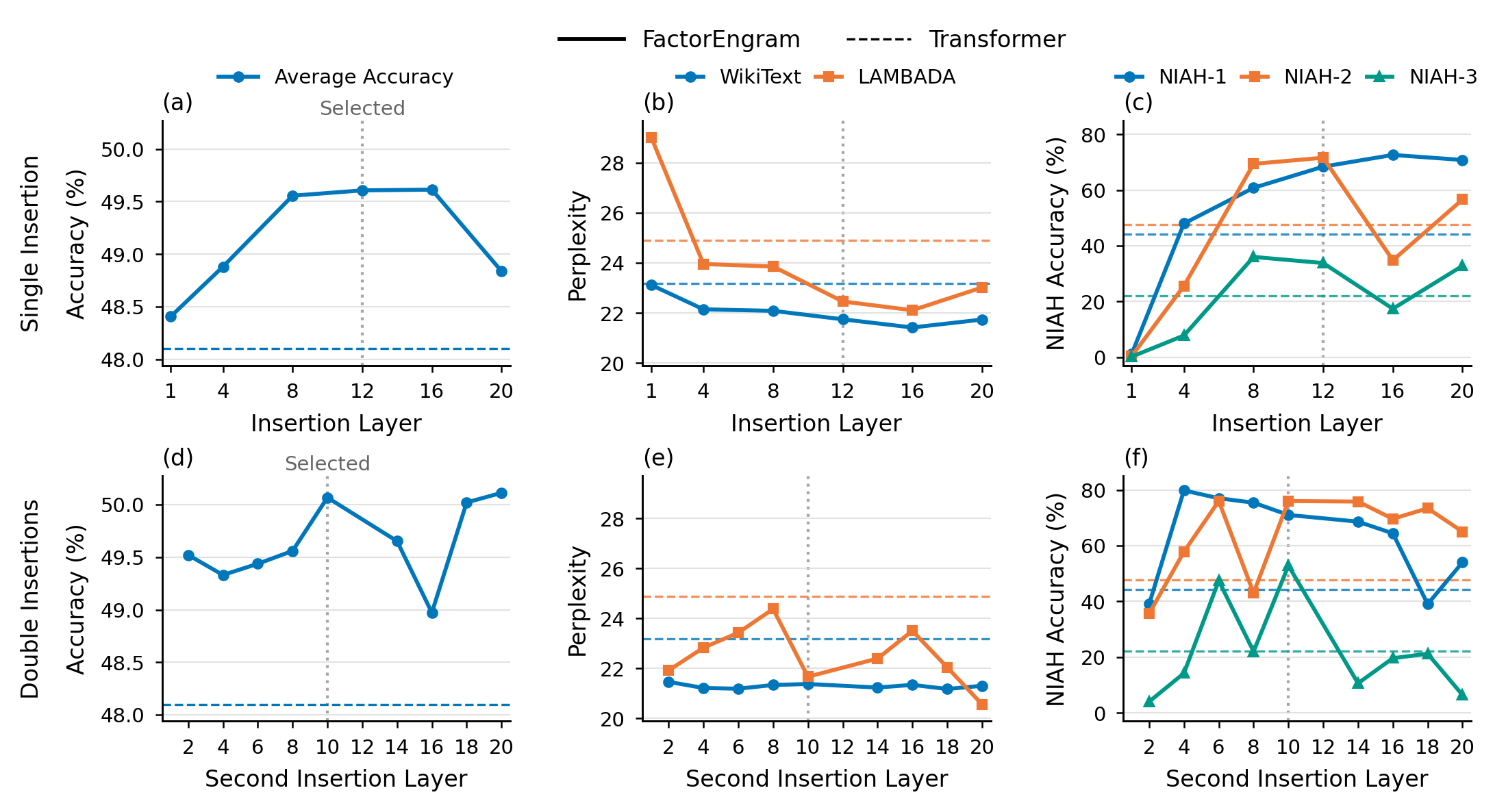}
\caption{Insertion-depth sweeps.
The top row varies the single insertion layer and
the bottom row fixes one insertion at optimal single layer 12 and varies the second. We report average downstream accuracy (\%), perplexity,
and NIAH accuracy (\%).
Colors distinguish metrics. Solid and dashed lines represent
FactorEngram and Transformer, respectively.
Dotted vertical lines mark the selected optimal depth.
Lower perplexity and higher accuracy are better.}
\label{fig:insertion-depth-sweep}
\vspace{-10pt}
\end{figure*}

We sweep the insertion depth of a single memory module,
then fix one module at layer 12 (optimal single-layer depth) and vary the second
module's depth (Figure~\ref{fig:insertion-depth-sweep}).
For single-layer insertion, moving from the earliest
layers toward the middle generally improves performance,
particularly on long-context retrieval.
Layer 12 offers a favorable balance across evaluations,
combining near-best down-stream and NIAH performances with the highest
NIAH-2 score, and is selected as the single-layer configuration.
With one module fixed at layer 12, sweeping the second
insertion depth produces substantial variation in NIAH scores.
The 10 \& 12 layer configuration achieves the highest NIAH-3 accuracy
among the tested pairs while maintaining strong performance
on the other evaluations.
We therefore use layers 10 and 12 as the default configuration.

\subsubsection{Within-Layer Placement}
\label{sec:insertion-locations}

We compare memory insertion before the attention module, before the
feed-forward network (FFN), and inside the SwiGLU~\citep{shazeer2020glu} FFN at its up-projection branch
(Table~\ref{tab:insertion-locations}).
Insertion before attention performs best overall,
achieving the highest on all accuracy metrics, as well as the lowest LAMBADA
perplexity.
Before-FFN insertion yields nearly the same WikiText
perplexity but substantially lower NIAH-3 accuracy of only 18.8, while inside-FFN insertion performs the
worst across all evaluation metrics.
We therefore adopt before-attention insertion as the
default configuration.

\begin{table*}[t]
\centering
\caption{Comparison of memory insertion locations with the 340M
backbone and an 8K context length. Accuracy metrics are reported as
percentages. Bold indicates the best value in each column.}
\label{tab:insertion-locations}
\resizebox{0.9\textwidth}{!}{%
\begin{tabular}{l|cc|c|ccc}
\toprule
\multirow{2}{*}{Insertion location}
& \multicolumn{2}{c|}{Language Modeling PPL}
& \multicolumn{1}{c|}{Downstream}
& \multicolumn{3}{c}{Long-context Retrieval} \\
\cmidrule(lr){2-3}
\cmidrule(lr){4-4}
\cmidrule(lr){5-7}
& WikiText $\downarrow$
& LAMBADA $\downarrow$
& Avg. Acc. $\uparrow$
& NIAH-1 $\uparrow$
& NIAH-2 $\uparrow$
& NIAH-3 $\uparrow$ \\
\midrule
Transformer & 23.19 & 24.90 & 48.10 & 44.2 & 47.7 & 22.1 \\
\rowcolor{gray!15}
Before attention & 21.29 & \textbf{21.69}
& \textbf{49.82} & \textbf{70.6} & \textbf{79.1} & \textbf{58.3} \\
Before FFN & \textbf{21.28} & 22.04 & 48.40
& 65.8 & 63.8 & 18.8 \\
Inside FFN & 23.34 & 26.91 & 47.90 & 47.4 & 47.4 & 12.2 \\
\bottomrule
\end{tabular}%
}
\vspace{-10pt}
\end{table*}

\section{Related Work}
\label{sec:related-work}

\textbf{Lookup-Based Memory for LLMs.}
Lookup-based memory architectures differ in representations,
contextual modulation, and integration with the backbone.
Engram~\citep{chengConditionalMemoryScalable2026} directly stores
N-gram embeddings and scales it with a scalar gate, without memorizing unigrams.
Gemma 3n's PLE~\citep{google2025gemma3nple} only memorizes single tokens and uses coordinate-wise gates computed from hidden
states alone.
STEM~\citep{sadhukhanSTEMScalingTransformers2026} highly relies on specific SwiGLU~\citep{shazeer2020glu} structure of FFN and replaces its
up-projection output with lookup embeddings only for token, whereas
FactorEngram preserves the original backbone model and adds memory as an auxiliary branch for longer N-gram patterns.
LongCat-Flash-Lite~\citep{liuScalingEmbeddingsOutperforms2026}
implements N-gram lookup memory but only as an augmentation for input embeddings in the main method.
Appendix~\ref{app:lookup-memory} maps the details of these architectures to
the operations defined in Section~\ref{sec:preliminaries}.

\textbf{Engram Variants and Extensions.}
Engram variants mainly modify memory representation and memory content coverage. Table~\ref{tab:engram-variants} in Appendix~\ref{app:engram-variants} summarizes the differences with FactorEngram in these two aspects. TN-gram~\citep{zhou2026tensorizing} also proposes a factorized variant,
which composes N-gram representations from shared token-position
factors but does not cover unigram and retains scalar gating.
Other variant primarily modify addressing, memory sources, or
storage. Lngram~\citep{zheng2026lngram} learns discrete keys from
hidden states. Engram-Nine~\citep{lin2026collisionfree} removes hash
collisions for frequent patterns. Tokenizer-Agnostic
Engram~\citep{lim2026tokenizeragnostic} uses byte-level rather than token-level hashing to map patterns to the same tables under tokenizer transfer.
Memory Grafting~\citep{cheng2026memorygrafting} retrieves frozen lookup table from another pretrained donor model with a trainable Engram fallback, while   
TF-Engram~\citep{ma2026tfengram} combines frozen phrase vectors
with memory hierarchy.
\citet{ma2026poolingengram} extends
memory storage and serving.
These directions are distinct from our improvements on factorized representation and modulation.

\textbf{Sparse Coding and Dictionary Learning.}
Sparse coding represents each signal with a small subset of components
from a shared dictionary~\citep{olshausenEmergenceSimplecellReceptive1996}.
This enables different signals to reuse the same components.
Sparse autoencoders apply this idea to activations of trained language
models and extract relatively independent features for interpretability~\citep{brickenMonosemanticityDecomposingLanguage2023,templeton2024scaling}.
To support local patterns to share memory components, FactorEngram stores
their coefficients in lookup tables and learns them jointly with a
shared dictionary and backbone under next-token prediction and an
$L_1$ penalty, instead of the activation reconstruction objective.

\section{Conclusion}
\label{sec:conclusion}


We introduced FactorEngram, a lookup-based memory architecture with factorized n-gram memory and basis-level contextual gating, in which local token patterns are represented by sparsity-regularized coefficients over a shared dictionary.
By reusing the dictionary for contextual gating
and reconstruction, FactorEngram allows patterns to share memory components
while modulating their contributions individually according to context.
Experiments show an overall improvement compared with the Transformer and the Engram baselines, with particularly
strong gains in long-context retrieval. 
A limitation of this study is that we evaluate FactorEngram only on 340M and 1B parameters. Although 1B-parameter models remain useful for edge computing and on-device deployment, future work will assess whether these benefits persist at larger scales.

\bibliography{references}
\bibliographystyle{colm2024_conference}

\clearpage
\appendix

\section{Detailed Comparison of Related Work}
\label{app:related-work}

\subsection{Lookup-Based Memory Architectures}
\label{app:lookup-memory}

Lookup-based memory architectures differ in how they represent local
patterns and integrate retrieved information into the backbone.
We describe representative architectures using the addressing, aggregation,
modulation, output mapping, and integration operations introduced in
Section~\ref{sec:preliminaries}.

\textbf{Engram.}
Engram~\citep{chengConditionalMemoryScalable2026} augments the backbone with a conditional
memory branch for local N-gram (length $n\in\{2,3\}$) patterns. It concatenates the retrieved dense vectors of each $\phi_b^{(\ell)}$
in aggregation $\mathcal{A}_{\ell}$. Contextual modulation
$\mathcal{C}_{\ell}$ uses a scalar gate derived from similarity between the hidden state
and a projected memory key to scale the retrieved content as a whole.
The value projection and short causal convolution form
$\mathcal{R}_{\ell}$, followed by residual integration through
$\mathcal{F}_{\ell}$. In contrast, FactorEngram additionally retrieves unigram memory
and represents the retrieved content as sparsity-regularized dictionary
coefficients, which are modulated for each individual dictionary basis instead of being scaled as a whole.

\textbf{Gemma PLE.}
Gemma 3n's Per-Layer Embeddings (PLE)~\citep{google2025gemma3nple} supply token-specific
representations at multiple layers.
For each single token, direct indexing retrieves layer-specific
vectors, which are combined with layer-specific projections of the
main input embeddings to form the memory input.
Its $\mathcal{C}_{\ell}$ applies a coordinate-wise gate computed
from the hidden state alone, without an inner product between
hidden-state and memory representations.

\textbf{STEM.}
STEM~\citep{sadhukhanSTEMScalingTransformers2026} incorporates token-specific memory
directly into the FFN by replacing the SwiGLU up-projection output. Direct token addressing retrieves an embedding vector, and
$\mathcal{A}_{\ell}$ is the identity for this single branch.
The retained SwiGLU gate applies coordinate-wise modulation through
$\mathcal{C}_{\ell}$, and the down-projection implements
$\mathcal{R}_{\ell}$. STEM's memory integration modifies the FFN computation and relies on the SwiGLU module,
whereas FactorEngram adds a memory branch while preserving the original
backbone architecture. Beyond this integration difference, FactorEngram uses sparsity-regularized coefficients and retrieves multi-token N-gram patterns.

\textbf{LongCat-Flash-Lite.}
LongCat-Flash-Lite also explores N-gram embedding expansion as a direction
for scaling model capacity~\citep{liuScalingEmbeddingsOutperforms2026}.
For its main architecture, we can express the operations in our notation as follows:
$\mathcal{A}_{\ell}$ concatenates the retrieved entries, contextual modulation $\mathcal{C}_{\ell}$ is identity, output mapping $\mathcal{R}_{\ell}$ performs the branch
projections and combination, and $\mathcal{F}_{\ell}$ integrates the result
before the first Transformer block.
The report also studies Per-Layer N-gram Embeddings (PLNE), which
introduce N-gram representations into the FFN using its
gating and down-projection.

\begin{table*}[t]
\centering
\caption{Comparison of Engram and its variants. Check marks indicate features
reported in each main method. Factorization refers to parameterizing memory entries
through shared factors. Fine-grained gating refers to assigning
component-wise contextual weights to a memory representation.
Sparse coding refers to explicitly encouraging sparsity within representations
through constraints or regularization. Memory content excludes backbone input
token embeddings. $^{*}$Lngram retrieves N-grams of learned discrete latent symbols
rather than input tokens.}
\label{tab:engram-variants}
\begingroup
\small
\setlength{\tabcolsep}{3pt}
\renewcommand{\arraystretch}{1.0}
\resizebox{\textwidth}{!}{%
\begin{tabular}{l|ccc|cc}
\toprule
\multirow{2}{*}{Method}
& \multicolumn{3}{c|}{Memory Representation}
& \multicolumn{2}{c}{Memory Content} \\
\cmidrule(lr){2-4}
\cmidrule(lr){5-6}
& Factorization
& Fine-Grained Gate
& Sparse Coding
& Unigram ($N=1$)
& N-gram ($N\geq 2$) \\
\midrule
Engram & & & & & $\checkmark$ \\
TN-gram & $\checkmark$ & & & & $\checkmark$ \\
Lngram & & & & & $\checkmark$\rlap{$^{*}$} \\
Engram-Nine & & & & & $\checkmark$ \\
Tokenizer-Agnostic Engram & & & & $\checkmark$ & $\checkmark$ \\
Memory Grafting & & & & & $\checkmark$ \\
TF-Engram & & & & & $\checkmark$ \\
CXL-based Engram Pool & & & & & $\checkmark$ \\
\midrule
FactorEngram & $\checkmark$ & $\checkmark$ & $\checkmark$
& $\checkmark$ & $\checkmark$ \\
\bottomrule
\end{tabular}%
}
\endgroup
\end{table*}

\subsection{Engram Variants and Extensions}
\label{app:engram-variants}

Table~\ref{tab:engram-variants} compares the Engram variants in Section~\ref{sec:related-work}, by their memory representations,
contextual gating, and retrieved pattern types with FactorEngram.

\section{Experiment Details}
\label{app:experiment-details}

\subsection{Model Configurations}
\label{app:model-configurations}

Table~\ref{tab:factorengram-configurations} summarizes the FactorEngram
configurations and training settings described in
Sections~\ref{sec:method} and~\ref{sec:experiments}.

\begin{table*}[t]
\centering
\caption{Default FactorEngram configurations and training settings at
the two backbone scales.}
\label{tab:factorengram-configurations}
\begingroup
\small
\setlength{\tabcolsep}{6pt}
\renewcommand{\arraystretch}{1.0}
\begin{tabular*}{0.75\textwidth}{l|cc}
\toprule
Configuration & 340M Backbone & 1B Backbone \\
\midrule
Total Params & 1.4B & 3.9B \\
Backbone Params & 340M & 1B \\
Total Tokens & 30B & 120B \\
\midrule
Layers & \multicolumn{2}{c}{24} \\
Sequence Length & \multicolumn{2}{c}{8192} \\
Vocab Size & \multicolumn{2}{c}{32K} \\
Hidden Dimension $d$ & 1024 & 2048 \\
\midrule
Training Dataset & \multicolumn{2}{c}{FineWeb-Edu} \\
Training Regime & \multicolumn{2}{c}{Jointly from scratch} \\
Batch Size & \multicolumn{2}{c}{64} \\
Optimizer & \multicolumn{2}{c}{AdamW} \\
Weight Decay & \multicolumn{2}{c}{0.01} \\
LR Scheduler & \multicolumn{2}{c}{Cosine Decay} \\
Training Steps & 57344 & 229320 \\
Warmup Steps & 1024 & 1911 \\
Peak Learning Rate & 1e-3 & 4e-4 \\
\midrule
Unigram Table Capacity (Entries) & \multicolumn{2}{c}{32K} \\
Bigram Table Capacity (Entries) & \multicolumn{2}{c}{250K} \\
Trigram Table Capacity (Entries) & 250K & 750K \\
FactorEngram Dim $d_m$ & 3072 & 3840 \\
FactorEngram Coefficient Width $s$ & 3072 & 3840 \\
FactorEngram Num Heads per N-Gram & \multicolumn{2}{c}{4} \\
FactorEngram Layers & \multicolumn{2}{c}{[10, 12]} \\
FactorEngram $N$-Gram Coverage & \multicolumn{2}{c}{[1, 2, 3]} \\
Within-Layer Placement & \multicolumn{2}{c}{Before attention} \\
Sparsity Strength $\lambda$ & \multicolumn{2}{c}{$10^{-3}$} \\
\bottomrule
\end{tabular*}
\endgroup
\end{table*}

\subsection{Main Result Details}
\label{app:main-result-detail}

Table~\ref{tab:main-results-all} provides the detailed
results corresponding to Table~\ref{tab:main-results},
including the individual downstream-task accuracies
summarized by the average accuracy in the main table.
Language-modeling and long-context retrieval results
are also included for completeness.

\begin{table*}[t]
\centering
\caption{
Detailed main results with an 8K context length. We additionally report every downstream task accuracy (\%).
}
\label{tab:main-results-all}

\small
\setlength{\tabcolsep}{6pt}
\renewcommand{\arraystretch}{1.05}

\begin{tabular}{@{}llccc@{}}
\toprule
\multicolumn{5}{c}{\textbf{340M backbone}} \\
\addlinespace[2pt]
Category & Metric & Transformer & Engram & FactorEngram \\
\midrule

\multirow{2}{*}{\shortstack[l]{Language\\Modeling}}
& WikiText PPL $\downarrow$
& 23.19 & 22.29 & \textbf{21.29} \\
& LAMBADA PPL $\downarrow$
& 24.90 & 23.94 & \textbf{21.69} \\

\midrule
\multirow{9}{*}{Downstream}
& LAMBADA $\uparrow$
& 37.32 & 38.15 & \textbf{39.30} \\
& PIQA $\uparrow$
& 67.22 & \textbf{68.23} & 68.06 \\
& HellaSwag $\uparrow$
& 43.05 & 44.00 & \textbf{46.14} \\
& WinoGrande $\uparrow$
& 51.95 & 52.72 & \textbf{53.51} \\
& ARC-Easy $\uparrow$
& 59.19 & 60.31 & \textbf{62.84} \\
& ARC-Challenge $\uparrow$
& 28.22 & 29.10 & \textbf{29.86} \\
& SocialIQA $\uparrow$
& 37.34 & 38.38 & \textbf{39.40} \\
& BoolQ $\uparrow$
& 60.54 & \textbf{61.62} & 59.42 \\
& \textbf{Average $\uparrow$}
& 48.10 & 49.06 & \textbf{49.82} \\

\midrule
\multirow{3}{*}{\shortstack[l]{Long-context\\Retrieval}}
& NIAH-1 $\uparrow$
& 44.2 & 69.8 & \textbf{70.6} \\
& NIAH-2 $\uparrow$
& 47.7 & 37.6 & \textbf{79.1} \\
& NIAH-3 $\uparrow$
& 22.1 & 15.0 & \textbf{58.3} \\

\midrule
\addlinespace[3pt]
\multicolumn{5}{c}{\textbf{1B backbone}} \\
\addlinespace[2pt]
Category & Metric & Transformer & Engram & FactorEngram \\
\midrule

\multirow{2}{*}{\shortstack[l]{Language\\Modeling}}
& WikiText PPL $\downarrow$
& \textbf{15.57} & 15.60 & \textbf{15.57} \\
& LAMBADA PPL $\downarrow$
& 11.73 & 11.83 & \textbf{11.39} \\

\midrule
\multirow{9}{*}{Downstream}
& LAMBADA $\uparrow$
& 47.65 & 47.62 & \textbf{48.85} \\
& PIQA $\uparrow$
& 72.42 & 72.31 & \textbf{72.46} \\
& HellaSwag $\uparrow$
& 56.76 & \textbf{57.46} & 57.02 \\
& WinoGrande $\uparrow$
& 58.20 & 58.41 & \textbf{58.48} \\
& ARC-Easy $\uparrow$
& 70.82 & \textbf{71.17} & 70.16 \\
& ARC-Challenge $\uparrow$
& 37.76 & \textbf{38.40} & 36.94 \\
& SocialIQA $\uparrow$
& 40.69 & 40.48 & \textbf{41.60} \\
& BoolQ $\uparrow$
& 60.58 & 60.63 & \textbf{60.88} \\
& \textbf{Average $\uparrow$}
& 55.61 & \textbf{55.81} & 55.80 \\

\midrule
\multirow{3}{*}{\shortstack[l]{Long-context\\Retrieval}}
& NIAH-1 $\uparrow$
& 47.5 & 45.8 & \textbf{72.0} \\
& NIAH-2 $\uparrow$
& 72.8 & 75.2 & \textbf{84.5} \\
& NIAH-3 $\uparrow$
& 25.7 & 16.6 & \textbf{44.2} \\

\bottomrule
\end{tabular}
\end{table*}

\end{document}